\documentclass[runningheads]{llncs} % for double space and review article
\usepackage{subfigure}

\usepackage{graphicx}
\usepackage{amsfonts}
\usepackage{hyperref}
\usepackage{amssymb}
\usepackage{mathtools}
\usepackage{xcolor,soul}
\usepackage{longtable}
\usepackage{ragged2e}
\usepackage{caption}
\usepackage{multirow}
\usepackage{amsmath}
\usepackage{tablefootnote}
\usepackage[ruled,vlined,linesnumbered]{algorithm2e}
\usepackage{float}
\usepackage{booktabs}
\usepackage{enumitem}

\begin{document}
	\author{Lalita Kumari\inst{1}, Sukhdeep Singh\inst{2}, Vaibhav Varish Singh Rathore\inst{3} \and
		Anuj Sharma\inst{1}}
	
	\institute{Department of Computer Science and Applications, Panjab University, India
		\email{\{lalita,anujs\}@pu.ac.in}\\
		\url{https://anuj-sharma.in}\and
		D.M. College(Aff. to Panjab University, Chandigarh), Moga, India\\
		\email{sukha13@ymail.com}\and
		Physical Research Laboratory, Ahmedabad, India\\
		\email{vaibhav@prl.res.in}}

		\title{A Comprehensive Handwritten Paragraph Text Recognition System: LexiconNet}
		\maketitle  
	
		\begin{abstract}
			\justifying
			In this study, we have presented an efficient procedure using two state-of-the-art approaches from the literature of handwritten text recognition as Vertical Attention Network and Word Beam Search. The attention module is responsible for internal line segmentation that consequently processes a page in a line-by-line manner. At the decoding step, we have added a connectionist temporal classification-based word beam search  decoder as a post-processing step. In this study, an end-to-end paragraph recognition system is presented with a lexicon decoder as a post-processing step. Our procedure reports state-of-the-art results on standard datasets. The reported character error rate is 3.24\% on the IAM dataset with 27.19\% improvement, 1.13\% on RIMES with 40.83\% improvement and 2.43\% on the READ-16 dataset with 32.31\% improvement from existing literature and the word error rate is 8.29\% on IAM dataset with 43.02\% improvement, 2.94\% on RIMES dataset with 56.25\% improvement and 7.35\% on READ-2016 dataset with 47.27\% improvement from the existing results. The character error rate and word error rate reported in this work surpass the results reported in the literature. 
			
			%		
			%		\begin{highlights}
				%			\item We proposed a unified end-to-end trainable HTR system blended with a lexicon decoder.
				%		 \item The proposed procedure adopts two state-of-the-art algorithms as Vertical Attention Network \cite{Coquenet2022}  for training and Word Beam Search \cite{Scheidl2018} in prediction stage. 
				%			\item Experiments performed on IAM, RIMES and READ-2016 datasets surpasses  the results of literature.
				%			\item  Detailed discussion of each step involved in a generic HTR process.
				%		\end{highlights}
			%		
			%		
			%		\begin{keyword}
				%				Paragraph Handwritten Text Recognition \sep Neural Network \sep Connectionist Temporal Classification \sep Word Beam Search \sep Segmentation Free \sep Optical Character Recognition 
				%			
				%		\end{keyword}		
			%\end{frontmatter}
		\end{abstract}
		\section{Introduction}\label{sec:introduction}
		A Handwritten Text Recognition (HTR) system  understands handwritten text written on digital surfaces, on paper or any other media. The HTR problem is complex enough to be widely studied by computer vision researchers across the community. In the present work, we proposed a tightly coupled HTR system constrained by lexicon and able to recognize the given paragraph-based text in an end-to-end manner without the need for any external segmentation step. The Hidden Markov Model (HMM) is a widely applied technique in an evolutionary period of HTR \cite{yusof2003}. The HMM based HTR models lack the use of contextual information because in large sequences of texts the HMM based system only focuses on the current time step as per Markovian assumption. \par The Recurrent Neural Network (RNN) based approaches are able to overcome these short coming of HMM. In recent years, due to advancements in computational capacity, deep learning-based models, especially Convolutional Neural Network (CNN), RNN and a combination of both  Convolutional Recurrent Neural Network (CRNN)  produces state-of-the-art accuracies for HTR task \cite{shi2017}. \par Although the deep learning based HTR methods give promising results, they require identification of the Region Of Interest (ROI), preprocessing and segmentation as a pre-step to recognize the given document. Hence, these latent errors restrict the system's accuracy and real-world applicability of the systems. In the present work, we aim to propose an end-to-end HTR system that can be as close to real-world applicability as possible. Thus, we are looking forward to developing a segmentation-free HTR system that can be trainable in an end-to-end manner confined by lexicon knowledge. Hence, by developing the system in such a manner, we are free from the latent errors and extra computational costs involved in segmentation followed by recognition approaches. In this study, we focus on offline HTR. In this paper, we presented an efficient and robust end-to-end paragraph based HTR system that takes a paragraph image as an input and generates its features in a line-by-line manner using a series of convolutional and depth-wise separable convolutional operations. These features are further sent to the decoder. The Word Beam Search (WBS) \cite{Scheidl2018} is used as decoder. We trained our NN model on line level first, then transferred these trained weights to the page level model for  training the whole page in an end-to-end manner. The following are the major contributions made by this study:
		\begin{itemize}
			\item We proposed an efficient procedure that unified end-to-end trainable HTR system blended with a lexicon decoder.
			\item The proposed procedure adopts two state-of-the-art algorithms as Vertical Attention Network \cite{Coquenet2022}  for training and the WBS \cite{Scheidl2018} in prediction stage.
			\item We have achieved state-of-the-art results by improving  Character Error Rate (CER) and Word Error Rate (WER) as 27.19\% and 43.02\% respectively on the IAM dataset, 40.83\% and 56.25\% on the RIMES dataset and 32.31\% and 47.27\% on READ-2016 dataset from existing results.
			
			\item Detailed discussion of each step involved in a generic HTR process. It will be helpful for future readers to understand the whole process in a systematic manner. 
			
		\end{itemize}
		The rest of the paper is organized as follows, the key contributions in the literature related to HTR systems are discussed in section 2. Section 3 explains the detailed system design of the HTR system. Section 4 includes the experimental setup and results. In section 5, a detailed discussion is presented on the lexicon decoder including best and worst case analysis. Section 6 draws the conclusion of the present study.

		\section{Related Works}	\label{sec:relatedworks}
		In this section, the state-of-the-art techniques of the HTR systems have been discussed in an evolutionary manner. We start the section by discussing character-based recognition, followed by word and line level recognition and conclude with a discussion of promising techniques of paragraph text recognition.
		
		\subsection{Character/Word Level Recognition} 
		In the early days, methods based on isolated characters were widely used for HTR tasks. In a segmentation-recognition approach, words are segmented into groups of characters using dynamic programming techniques and recognised further using probabilistic models \cite{Bozinovic1989}. The isolated characters are obtained using various image processing techniques \cite{Seni1994}. Further, the adaptive word recognition method is used  with hand-crafted features \cite{Park2002}. While hand-crafted features are used along with HMM based models to recognize the words of large English vocabulary \cite{Yacoubi99}. In  a similar study, the input to the system is given by words and these  words are segmented into characters for further processing by the HMM \cite{kundu1988}.  Further, a combination of HMM + RNN is used to recognize at word level \cite{Senior1998}.
		
		\subsection{Line Level Recognition}
		Initial approaches to solving line level recognition are based on HMM. Statistical language model based techniques are used to recognize complete lines \cite{Bunke2004}. Further, sliding window-based features are extracted, and a given text line is converted into a sequence of temporal features using HMM \cite{ploetz2009}. As per Markovian assumption, each observation depends upon the current time step. Hence, context-based information is not fully utilized in HMM based models. In later years, CNN based system is a state-of-the-art method to extract the features of images and RNN is for remembering them for long; hence the combination of both systems, CRNN, emerged as a new state-of-the-art approach. In one such study, line recognition is done using CTC with BLSTM along with a token passing algorithm \cite{Graves2009}. Further, the combination of  CNN with MDLSTM and BLSTM  is used to recognise the handwritten text \cite{Moysset2017}\cite{Wigington2017}. The CRNN architecture is a combination of deep CNN and RNN layers and is able to produce state-of-the-art results in recognizing sequential objects in images in scene text recognition\cite{shi2017}, which is further extended in HTR domain \cite{SimpleHtr2018}. The total number of trainable parameters can be further reduced using Gated CNN based NN models. In one such study, a system  based upon Gated CNN and bidirectional gated recurrent units are able to produce state-of-the-art results \cite{flor2020}. In a similar study, NN based language model is integrated in the decoding step of HTR. In this study,  different type of HTR architecture is examined in connection with NNLM at the decoding step on standard datasets \cite{zamora2014lm}.
		
		\subsection{Paragraph Level  Recognition}
		There are two types of approaches followed in literature. First, the segmentation techniques  are used to obtain line level images of paragraph \cite{Vassilis2010}. In one such segmentation technique, there has been a method proposed for word segmentation which is based upon identifying the connecting region \cite{manmatha1999}. The same algorithm is extended for line segmentation \cite{manmatha2005}. The statistical methods based segmentation techniques are also  used to  follow the cursive path of one's handwriting \cite{Arivazhagan2007}. After obtaining line images, line level optical recognition models are used to obtain the transcriptions of the same. But these segmentation based techniques induced error which resulted in poor recognition results. In the second method, recognition of the paragraph is done without any external segmentation. In these paragraph recognition techniques, either attention based techniques are used to do internal segmentation or the multi-dimensionality of the task is considered to achieve paragraph recognition in a single step. Following the first approach of segmentation free techniques, CNN along with BLSTM is also used to recognise text at paragraph level as an alternative to MDLSTM \cite{puigcerver2017}. In the second approach of paragraph recognition, the two dimensional aspect of the task is considered. It uses a Multi-Dimensional Connectionist Classification (MDCC) technique to process two dimensional features of paragraph images. By using a Conditional Random Field (CRF), a 2D CRF is obtained from the ground truth text. This CRF graph is the control block for line selection among multiple lines \cite{Schall2018}. In one similar study, it considered the paragraph in one single large text line. In this work, bi-linear interpolation layers are combined with feature extraction layers of the Fully Convolutional Network (FCN) encoder to obtain a large single line. The need for line breaks in transcription and line level pre-training is not required in this module\cite {yousef2020}. We have extended the study of Vertical Attention Network (VAN) \cite{Coquenet2022} by applying the WBS \cite{Scheidl2018} decoder as a post processing step, further improving the accuracies and contributing to the solution of the cursive HTR problem.
		
		\section{System Design}	\label{sec:architecture}
		\begin{figure}[!hbt]
			\captionsetup{justification=centering}
			\includegraphics [width=\linewidth, height=14cm]{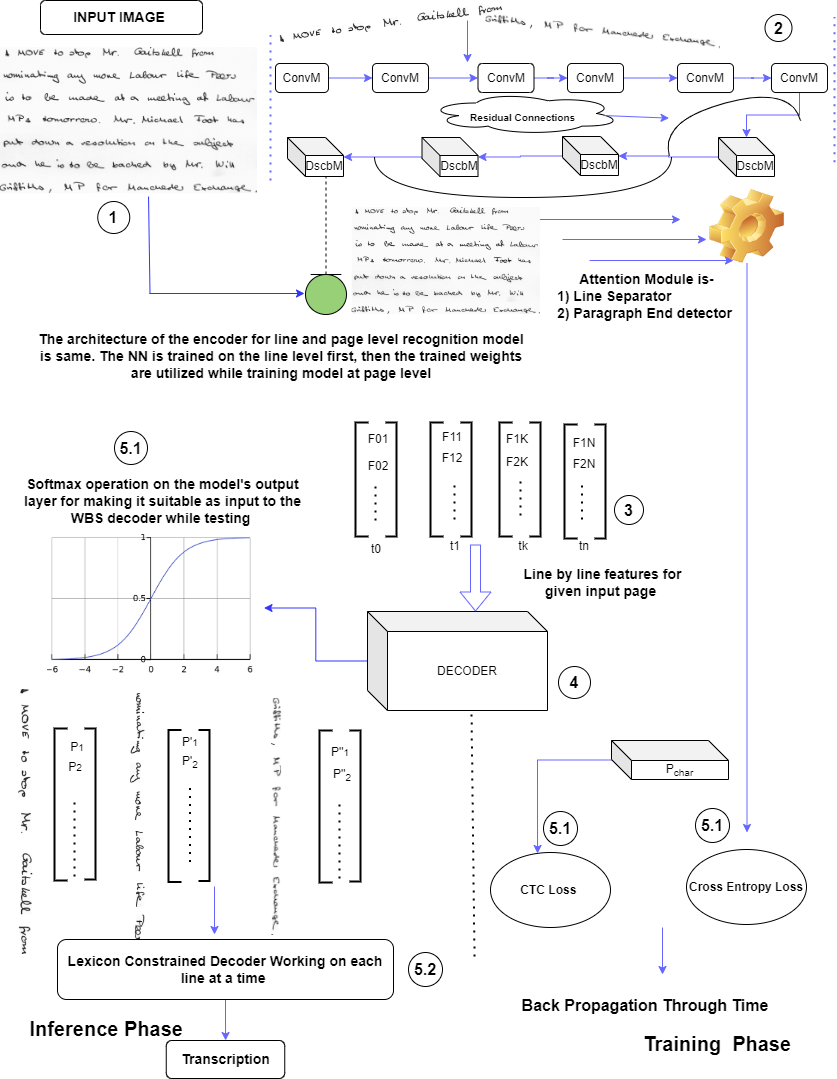}
			\caption{A Comprehensive Handwritten Paragraph Text Recognition System: LexiconNet}
			\label{figure:fig-1}
		\end{figure}
		
		\begin{figure}
			
			\includegraphics [width=\linewidth, height=7cm]{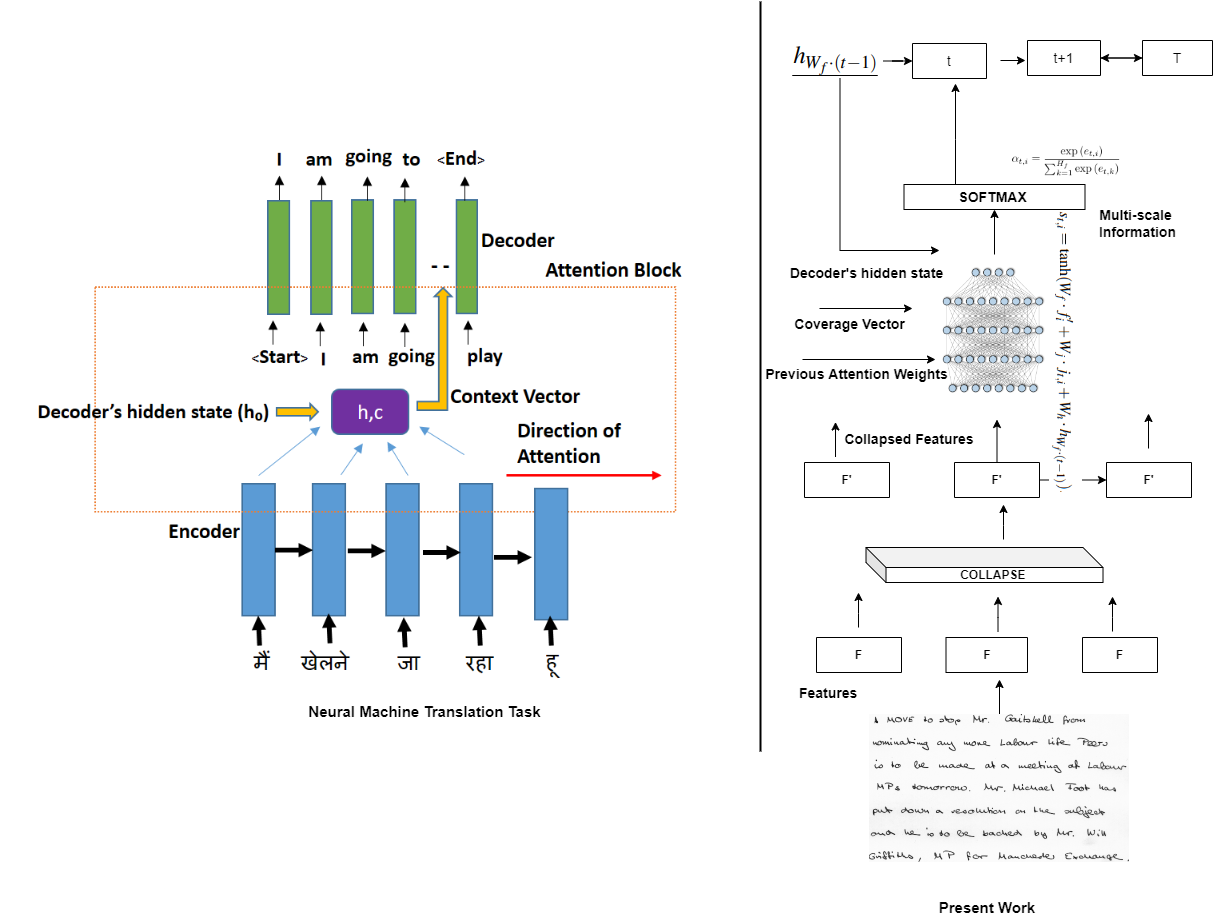}
			\caption {Comparison of Attention mechanism in Neural Machine Translation (NMT) task (left) and in the present study (right). 
			}
			\label{figure:fig-5}	
		\end{figure}
		
		In this study, we have performed end-to-end paragraph text recognition on IAM, RIMES and READ-2016 datasets. In this section, we have described the overall architecture in modular form. Module 1 is the recognition module which carries a VAN \cite{Coquenet2022}, and module 2 is the decoder module \cite{Scheidl2018}. In figure \ref{figure:fig-1}, we have shown the functionality of both modules together. In this figure, ConvM denotes a single block of convolution operation of Module 1 and DscbM denotes a single block of depthwise separable operation of Module 1. Initially, in step 1, the input image is preprocessed. In parallel to step 1, the line level model of IAM, RIMES and READ-16 datasets are trained and their weights are used to train the paragraph level model of the same dataset in an end-to-end manner.  The attention mechanism is used to obtain lines from paragraphs as step 3. In step 4, the LSTM decoder is used in combination with step 3 to obtain the occurrence of each character. In step 5, depending upon the processing stage, we either backpropagate through time or apply the WBS decoding algorithm. We have used all the model hyperparameters  of training and testing, the same as mentioned in \cite{Coquenet2022}. Module 1 and Module 2 are explained as follows, 
		\par Module-1 consists of a feature extraction module, internal line segmentation  and end-of-paragraph detection sub-modules. As the name suggested, this module takes a greyscale or colour input image and produces its feature map. This module consists of Convolution and Depthwise separable Convolutional blocks. These blocks magnify the context. Composite dropout (standard  with 0.5 and 2D with 0.25 probability at specified location) is used in this system. The Convolution block  has two convolution layers followed by the normalization layer and convolutional layer. The convolution operations are the building block of a convolutional layer. In convolution operation, a kernel slides over the input image and produces feature maps of images. A (3×3) kernel followed by Rectified Linear Unit (ReLU) activation function has been used in this module. A convolution converts all the pixels of the receptive field (961 × 337) into a single value. Thus, image size is reduced and brings the receptive field’s collective information together. In this work, height is reduced by a factor of 32 and width is reduced by a factor of 8 which is    controlled by the stride size of the last convolutional layer of each block. The depthwise separable convolution blocks  are similar to convolution blocks. In this, instead of using a standard convolutional layer,  depthwise separable convolutions are used. The depthwise separable convolutions are computationally less costly than regular convolutions with the same level of performance. The last convolutional layer of the DscbM has a fixed stride of (1, 1) to preserve the input shape. Residual connections are used to address the vanishing gradient problem in deeper NN architectures. In this,  residual connections with the element-wise sum operator after the last convolutional layer of the final convolution block  and each depthwise separable convolution  block are used  to propagate the weight updates up to the first layer while training without changing the output. In the Internal line segmentation module, each paragraph image is internally segmented into lines and generates its feature at each time step using Bahdanau soft attention \cite{bahdanau2014} mechanism. The attention is used to identify a particular line of a given page at a specific time step. If we have L lines in a given page so by using the attention mechanism i\textsuperscript{th} line features will be calculated at i\textsuperscript{th} time step. The probabilistic distribution of these attention weights gives the importance of each frame at time t. The attention is applied along the vertical axes such that it only needs one high attention weight value to identify a line, the rest of the attention weights can be near zero. Figure  \ref{figure:fig-5} summarizes the attention process used in the present study and compares it with attention used in NMT task. In NMT, each word like 
		\lq Main\rq ,\ \lq Khelne\rq \  has different importance at different timestep of translation and the attention weights are used to signify this. Thus attention is applied horizontally in it. While in the present setup, attention is used to identify features in a linewise manner from a given paragraph image. Thus attention applies vertically.  The task of paragraph end detection is separately taken as a classification task with two classes, one class represents the end of the paragraph while the other represents that paragraph has not ended yet. To identify it, we leverage the use of contextual information  we obtained from the attention step and the decoder’s hidden state information. We have obtained the information about the total number of lines contained by each page while processing the dataset. This serves as one hot encoded ground truth information. The Cross-Entropy (CE) loss function for training purposes, gives us proximity of our predictions to the ground truth and updates the weights accordingly.  Thus total loss for training will be the weighted  sum of CTC loss and CE loss. Since both the losses are the same equal weights are considered while summation.  The LSTM is used as a decoder block. Each decoder's current step is input by the last step’s hidden state and current line features. A single LSTM layer is used with 256 cells in the decoder module. The initial hidden state (h0) of LSTM for the first line of the page is initialized with zeros. Hidden states are kept preserved from one line to another line in a paragraph to use contextual information of the paragraph. LSTM’s output tensor dimension is further extended to apply convolution with kernel size equal to 1 and input channel size is same as cells in the LSTM layer, and output channel size equal to the total number of character set in dataset +1 (for CTC Blank) \cite{Coquenet2022}.
		\begin{figure}[t]
			\includegraphics [width=\linewidth, height=5cm]{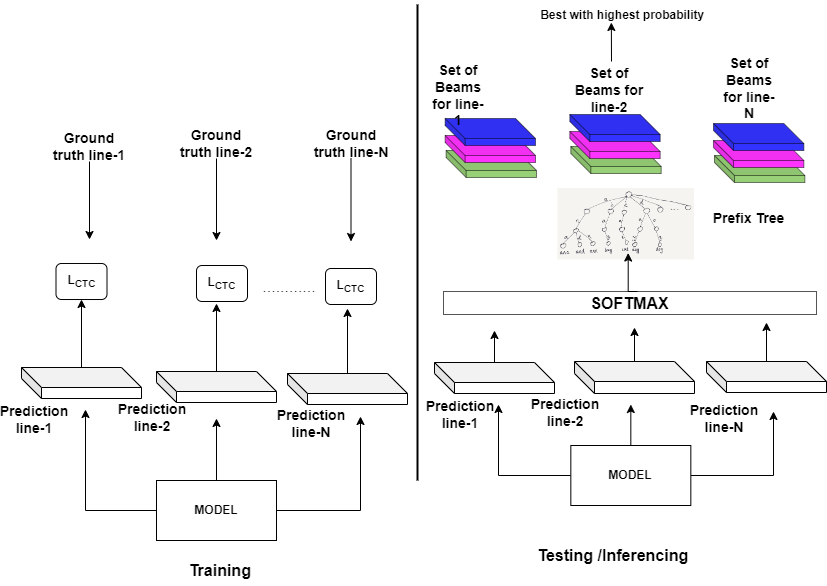}
			\caption {Various roles of decoder}	
			\label{figure:fig-6}
		\end{figure}
		
		\par Module-2 is used to predict the final sequence of the text at the testing stage. In this work, we have used the WBS decoding algorithm instead of the best path decoding algorithm.  Figure \ref{figure:fig-6} explains the role of the decoder while training and testing. During the training phase, CTC loss is calculated against the predicted, and the actual text and model weights are updated  using back propagation accordingly. As in Figure \ref{figure:fig-6} for each line feature  at each time step,  we calculate the CTC loss and update the weights. While in testing, the decoder’s output is softmax to make it suitable for this module.  The decoding algorithm starts with an empty beam. A beam is a notion of a possible transcription candidate at a particular time step. A prefix tree is also built with the available corpus. At t=0, this prefix tree is queried for all possible words from the corpus and beam extended in the same manner. From t=1 to T, where T is the total time steps, beam extension depends upon the state of the beam. A beam can be either in a word state or in a non-word state. A beam is transitioned to non-word stage when non-word characters are added to it. Only the \lq word\rq \  state beam is extended by the characters, that are given by the prefix tree containing the prefix, while in \lq non-word\rq \  state beam can be extended using all the words as well as non-words characters. We have used the \lq Words\rq \  operation mode of the decoding algorithm with a beam width of 50.

		\section{Experimental Setup and Results}
		\vspace{-1em}
		In this section, we have covered the experimental setup of the present study. It includes datasets used, preprocessing methods applied, evaluation metrics and training details. We have used the python programming language along with PyTorch framework with the apex to implement the present work. We have implemented  CTC WBS decoder\footnote[1]{https://github.com/githubharald/CTCWordBeamSearch} with basic architecture of  VAN\footnote[2]{https://github.com/FactoDeepLearning/VerticalAttentionOCR/}.
		\vspace{-1em}
		\subsection{Datasets}
		The architecture presented in the paper is evaluated against IAM \cite{IAM}, RIMES \cite{rimes} and the READ-2016 \cite{read} datasets which are discussed as follows,
		
		\subsubsection{IAM}
		The IAM handwriting dataset includes the forms of handwritten text, which can be used for training and testing purposes of text recognition experiments. We have used paragraph and line level segmentation in this work with the train, test and validation split are mentioned in Table \ref{tab:dataset}. 
	
		\subsubsection{RIMES}
		The RIMES dataset is a collection of handwritten letters sent by individuals to company administrators. Train, validation and test splits are shown in Table \ref{tab:dataset}.
		
		\subsubsection{READ-16}
		It is presented in ICFHR 2016 \cite{read}. It is written in the German language. Page, Paragraph and Line level segmentation is provided by READ-2016. The training, validation and testing splits are shown in Table \ref{tab:dataset}. 
		\vspace{-1em}	
		\begin{table}[!hbt]
			\centering
			\scriptsize
			\caption{Datsets splits in Training, Validation and Testing sets for present study}
			\begin{tabular}{lllll}
				Dataset	&      & Train & Validation & Test \\ \hline
				\multirow{2}{*}{\begin{tabular}[c]{@{}l@{}}IAM\\ (No. of Characters-79)\end{tabular}} & Line &6,482&976& 2,915\\ \cline{2-5} 
				& Page & 747 &116&336      \\ \hline
				\multirow{2}{*}{\begin{tabular}[c]{@{}l@{}}RIMES\\ (No. of Characters-100)\end{tabular}} & Line & 10,532 &801 &  778 \\ \cline{2-5} 
				& Page &  1400 & 100&100 \\ \hline
				\multirow{2}{*}{\begin{tabular}[c]{@{}l@{}}READ-2016\\ (No. of Characters-89)\end{tabular}}            & Line & 8,349 &1,040 & 1,138 \\ \cline{2-5} 
				& Page &   1,584  & 179&  197    \\ \hline
			\end{tabular}
			
			\label{tab:dataset}
		\end{table}
		\subsection{Preprocessing and Training Details}
			\begin{algorithm}			
			\SetAlgoLined
			\KwIn{Paragraph batch images \textit{I}, and ground Truth G\textsubscript{t} with lines {G\textsubscript{t1}}, {G\textsubscript{t2}}...{G\textsubscript{tl}} and text corpus Corp\textsubscript{text}}
			\KwResult{Training using backpropogation and Evaluation Metrics on given Dataset}
			Initialize main()\;
			Params=initParams()\;
			setDevice(Params)\;
			D\textsubscript{dataset}=loadDataset(Params)\;
			I=preprocessDataset(D\textsubscript{dataset})\;
			Training the model in an end-to-end manner \cite{Coquenet2022}\;
			pred\textsubscript{t}=Testing()\;
			pred\textsubscript{t}=Softmax(pred\textsubscript{t})\;
			Transcript=Concat(Transcript, WBS(pred\textsubscript{t}, Corp\textsubscript{text})) \cite{Scheidl2018}\;
			CER,WER= Accuracy(Transcript,ground truth)	
			\caption{End-to-End HTR Training and Testing}
			\label{algorithm:algo}	
		\end{algorithm}
		In this work, the  methods that have been applied to each dataset for preprocessing are the same. The  variable-sized images of datasets have been resized to a fixed size. Images are zero-padded to 800 pixels width and 480 pixels height to ensure minimum feature width of 100 and height of 15. Data augmentation techniques are used to increase variations which is available during training. There were no data augmentation techniques applied during testing. Change in resolution, perspective transformation, elastic distortions, random perspective transformation \cite{yousef2020}, dilation and erosion, brightness and contrast adjustment are  applied  with a probability of 0.2 in the given order same as \cite{Coquenet2022}.
		Algorithm \ref{algorithm:algo}, of the HTR system takes input from a set of images, a set of ground truth values and a text corpus. It produces CER and WER of the prediction of handwritten text as output. Algorithm \ref{algorithm:algo} is explained in this section in line by line manner as follows,
		
		\begin{itemize}
			\item[\textbf{Line 1-4:}]- Device setup in GPU/CPU mode (setDevice()), loading of dataset and set model parameters (loadDataset() and initParams()).\\
			
			\item[\textbf{Line 5:-}]  Pre-process the images (processedDataset()) and load train, test and validation batches (loadBatch()).\\
		
			\item[\textbf{Line 6:-}]  End-to-end training of the HTR model. CE loss and CTC loss are used to train the model. The attention mechanism helped in internal line segmentation. A pretrained model at line level on the same dataset is also used in the training for quick convergence.\\
			
			\item [\textbf{Line 7:-}]   pred\textsubscript{t} is the line-by-line character occurrence matrix for a given set of images at the testing stage.\\
			
			\item [\textbf{Line 8-9:-}] The probability matrix is given to WBS algorithm \cite{Scheidl2018}. This returns the recognized text of text line of the paragraph image.  By concatenating the WBS decoder's output of each line of the paragraph we have obtained the predicted text of the whole paragraph.\\
			
			\item[\textbf{Line 10:-}]Calculate accuracy of the system.
		\end{itemize}
		\subsection{Results and Comparision}
		\begin{table}[!htbp]
			\caption{Results and Comparision Table}
			\scriptsize
			\centering
			\begin{tabular}{cllll}
				\hline
				\textbf{Reference}       & \multicolumn{1}{c}{\textbf{\begin{tabular}[c]{@{}c@{}}CER(\%)\\ Valid\end{tabular}}} & \multicolumn{1}{c}{\textbf{\begin{tabular}[c]{@{}c@{}}WER(\%)\\ Valid\end{tabular}}} & \multicolumn{1}{c}{\textbf{\begin{tabular}[c]{@{}c@{}}CER(\%)\\ Test\end{tabular}}} & \multicolumn{1}{c}{\textbf{\begin{tabular}[c]{@{}c@{}}WER(\%)\\ Test\end{tabular}}} \\ \hline
				\multicolumn{5}{c}{\textbf{\quad \quad IAM Dataset}}                                                                                                                                                                                                                                                                                                                                           \\ \hline
				\multicolumn{1}{l}{CNN+MDLSTM\cite{bluche2017}} & --                                                                                 & --                                                                                    & 16.2                                                                                & --                                                                               \\ \hline
				\multicolumn{1}{l}{CNN+MDLSTM\cite{bluche2016}} & 4.9                                                                                & 17.1                                                                                   & 7.9                                                                                & 24.6                                                                              \\ \hline
				\multicolumn{1}{l}{RPN+CNN+BLSTM+LM\cite{curtis2018}} & --                                                                               & --                                                                                   & 6.4                                                                                & 23.2                                                                              \\ \hline

				\multicolumn{1}{l}{VAN\cite{Coquenet2022}} & 3.02                                                                                  & 10.34                                                                                    & 4.45                                                                                & 14.55                                                                               \\ \hline
				\multicolumn{1}{l}{Ours (LexiconNet)} & 1.89                                                                                    & 5.17                                                                                    & \textbf{3.24}                                                                                 & \textbf{8.29 }                                                                             \\ \hline
				
				\multicolumn{5}{c}{\textbf{\quad \quad RIMES Dataset}}                                                                                                                                                                                                                                                                                                                                         \\ \hline
				\multicolumn{1}{l}{CNN+MDLSTM\cite{bluche2016}}     &  2.5                                                                                    &  12.0                                                                                    &   2.9                                                                                  &12.6                                                                                     \\ \hline
				\multicolumn{1}{l}{RPN+CNN+BLSTM+LM\cite{curtis2018}} & --                                                                               & --                                                                                   & 2.1                                                                                & 9.3                                                                              \\ \hline
				\multicolumn{1}{l}{VAN\cite{Coquenet2022}} & 1.83                                                                                  & 6.26                                                                                    & 1.91                                                                                & 6.72                                                                               \\ \hline
				\multicolumn{1}{l}{Ours (LexiconNet)} & 1.06                                                                                   & 2.91                                                                                    & \textbf{1.13}                                                                                 & \textbf{2.94}                                                                             \\ \hline
				
				\multicolumn{5}{c}{\textbf{\quad \quad READ-2016 Dataset}}                                                                                                                                                                                                                                                                                                                                     \\ \hline
				\multicolumn{1}{l}{VAN\cite{Coquenet2022}} & 3.71                                                                                  & 15.47                                                                                    & 3.59                                                                                & 13.94                                                                               \\ \hline
				\multicolumn{1}{l}{Ours (LexiconNet)} & 2.29                                                                                 & 7.46                                                                                    & \textbf{2.43}                                                                                 & \textbf{7.35}                                                                            \\ \hline
				
			\end{tabular}
			\label{table:tab-3}	
		\end{table}
		
		In this section, an end-to-end paragraph recognition architecture LexiconNet is evaluated. We compare our results with other works. In decoder\cite{Scheidl2018}, we have used the 'Words' mode with beam width 50 and used the corpus present in the test dataset and validation  dataset respectively for test and validation error. Table \ref{table:tab-3} summarizes our findings as we have obtained 3.24\% CER and 8.29\% WER on the test dataset of IAM with 27.19\% improvement in CER and 43.02\% improvement in WER, 1.13\% CER and 2.94\% WER on the test dataset of RIMES with 40.83\% improvement in CER and 56.25\% improvement in WER and 2.43\% CER and 7.35\% WER on the test dataset of READ-2016 with 32.31\% improvement in CER and 47.27\% improvement in WER, respectively.  
		
		\subsection{Evaluation Metric}
		State-of-the-art evaluation metrics such as CER and WER are used  to evaluate the present study. The CER is based on Levenshtein Distance (\textit{$L\textsubscript{d}$}) between the ground truth (\textit{$g$}) and predicted text (\textit{$p$}). The WER is the same as CER, but it is evaluated on word level rather than character level as in equation (\ref{eq:12}) where $p_{sub}$ is the number of substitutions, $p_{ins}$ is the number of insertions and $p_{del}$ is the number of deletions with reference to the predicted text. $g_{total}$ is the total number of characters in the actual string.
		
		\begin{equation}\label{eq:12}
			\text {CER }=(p_{sub} + p_{ins}+ p_{del}) / g_{total}
		\end{equation}
		\vspace{-3em}	
		
		\section{Discussion}
		
		In the present study, the proposed end-to-end handwritten paragraph text recognition architecture surpasses the results reported in the literature. An HTR, architecture involves preprocessing, training and  post processing modules. Since, various heterogeneous modules are working in abstraction, comparing one deep NN architecture to another is not viable. We have added the  WBS  decoder \cite{Scheidl2018} as a post processing  step and extension to the VAN study \cite{Coquenet2022}.
		
		\subsection{Improvement in accuracy upon applying lexicon decoder in HTR System}
		
		\begin{figure}
			\centering
			\subfigure[]{\includegraphics[width=0.4\textwidth, height=6cm]{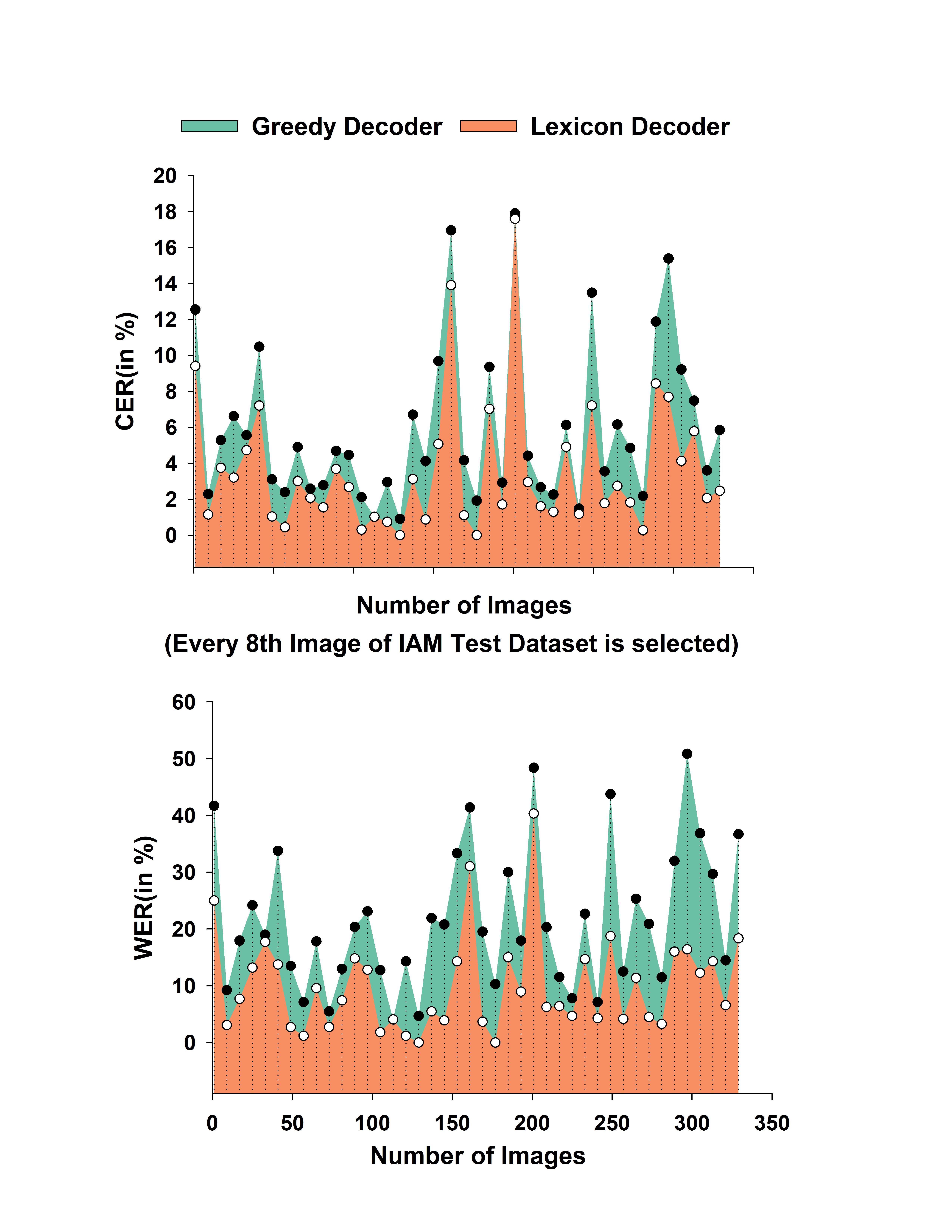}} 
			\subfigure[]{\includegraphics[width=0.4\textwidth, height=6cm]{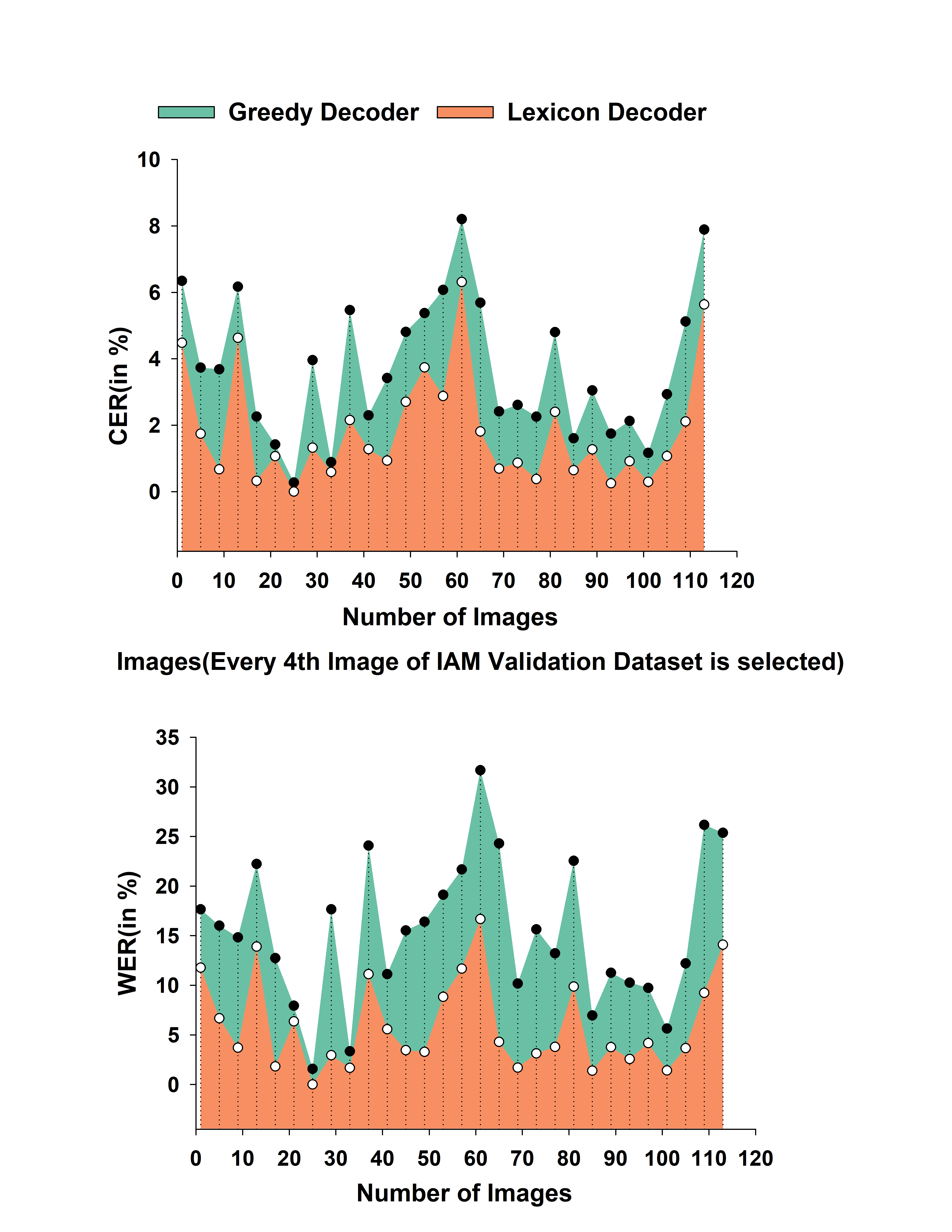}} 
			\caption{ Comparision of decoder(a) Test dataset (b) Validation dataset.}
			\label{figure:fig-7-8}
		\end{figure}
		
		We have summarized the effective percentage gain in CER and WER in 
		the present study in comparison with the existing one. As shown in Figure \ref{figure:fig-7-8} (a) and Figure \ref{figure:fig-7-8} (b) comparison of greedy decoder accuracy with lexicon decoder on IAM test dataset and validation dataset respectively. The graph shows that the rate of improvement of WER is more than CER. The decoding algorithm favours WER more.
		equation (\ref{eq:13}) and equation (\ref{eq:14})  are used to calculate the percentage of improvement in CER/WER. \textit{y\textsubscript{cer}} is the CER obtained from greedy decoder while $\hat{y}$\textsubscript{cer} is CER obtained from WBS decoder and \textit{y\textsubscript{wer}} is the WER obtained from greedy decoder while $\hat{y}$\textsubscript{wer} is WER obtained from WBS decoder.
		\begin{equation}\label{eq:13}
			\%  \text{ Improvement\textsubscript{cer}}=	\frac{y\textsubscript{cer}-\hat{y}\textsubscript{cer}}{y\textsubscript{cer}} *(100)
		\end{equation}
		\begin{equation} \label{eq:14}
			\%   \text{ Improvement\textsubscript{wer}}=\frac{y\textsubscript{wer}-\hat{y}\textsubscript{wer}}{y\textsubscript{wer}} *(100)
		\end{equation}
		\vspace{-2em}
		\subsection{Effect of Corpus Size} 
\begin{figure}[!hbt]
			\includegraphics [width=\linewidth, height=6cm]{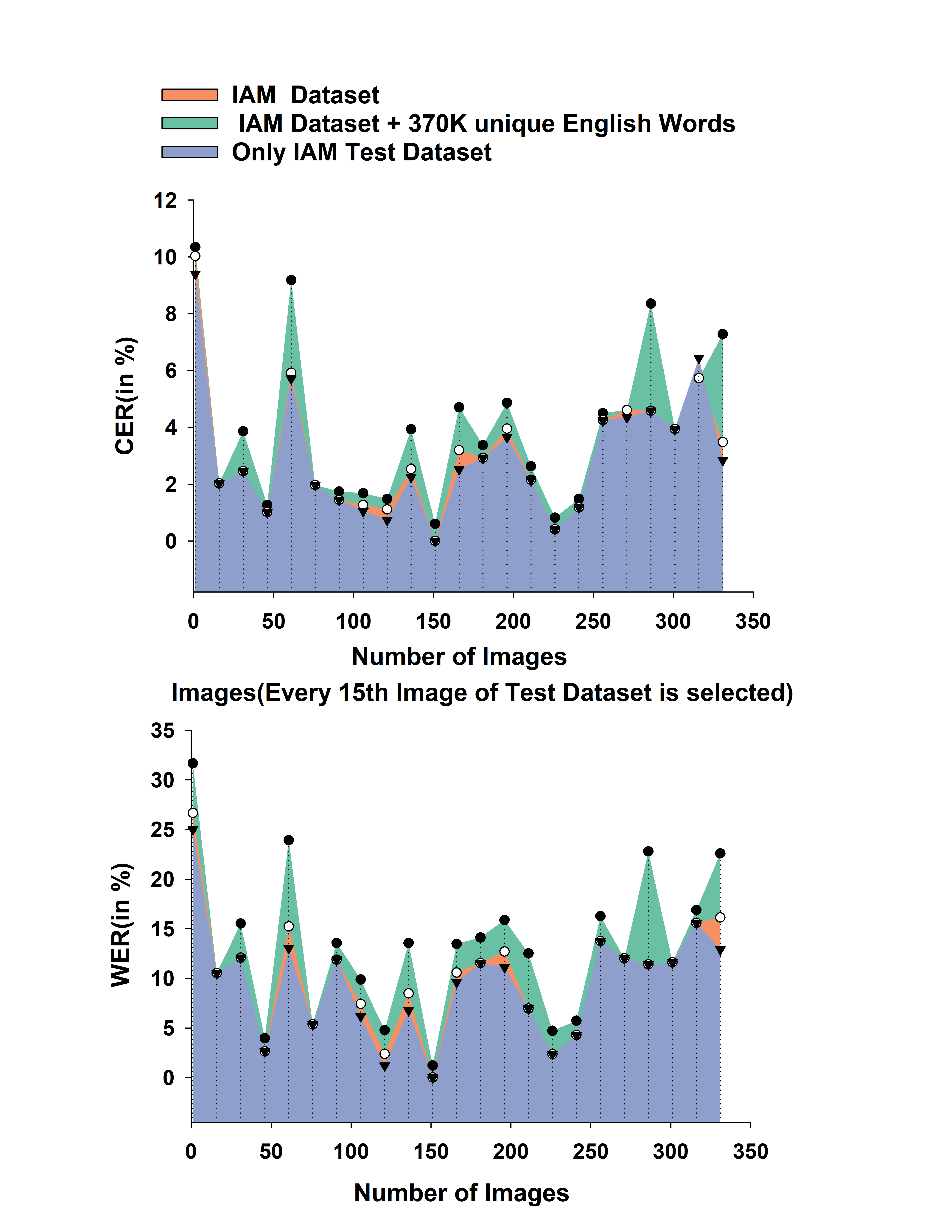}
			\caption {Grpah between CER and WER of various corpus sizes on Test dataset images of IAM dataset. It is also evident from the graph that, the choice of corpus affects the overall accuracy}
			\label{figure:fig-12}	
		\end{figure}

		\vspace{-3em}
		\begin{table}[!hbt]
			\caption{CER and WER value is various corpus sizes of IAM dataset}
			\scriptsize
			\centering
			\begin{tabular}{llcc}
				\toprule
				\multicolumn{1}{c}{\textbf{S.No.}} & \multicolumn{1}{l}{\textbf{Corpus Size}}                    & \textbf{\begin{tabular}[c]{@{}c@{}}CER\\ (In \%)\end{tabular}} & \textbf{\begin{tabular}[c]{@{}c@{}}WER\\ (in \%)\end{tabular}} \\ \toprule
				1                                & Only Test Dataset Corpus with Lexicon Decoder (LD)      & 3.24                                                             & 8.29                                                           \\ \hline
				2                                & IAM corpus with LD                     & 3.42                                                             & 9.02                                                            \\ \hline
				3                                & IAM corpus + 370K unique  words with LD & 4.19                                                             & 12.19                                                             \\ \bottomrule
			\end{tabular}
			\label{table:tab4}	
		\end{table}
		The scenario presented is not always favourable. Internally lexicon decoder creates a prefix tree containing all the possible prefix paths of words of a given dataset corpus for its processing and final transcription containing words of these paths. Hence, the accuracy also depends upon the type of the corpus used. In the present work, we have considered a corpus containing the text of images of the test dataset while testing the NN model and the text of images of the validation dataset while validating the model. As a part of the discussion, we considered different corpus sizes to estimate algorithm performance. In the first case, we consider all the text of the IAM dataset while in the second case we have added 370K unique English language words\footnote[3]{https://github.com/dwyl/english-words}  along with the whole IAM corpus. As per the results shown in Table \ref{table:tab4} and Figure \ref{figure:fig-12} it can be evident that, if we increase the size of the corpus, the total number of paths also increases. Thus the possibility of getting on the wrong path increases. Thus, our CER and WER values increase as we increase the corpus size. Table \ref{table:tab4} also shows that there is less increase in error while increasing the corpus size. Hence, the decoding algorithm is robust enough to handle a large corpus.
		
		\vspace{-2em}
		\subsection{Failure and Best Case Analysis}
		\begin{figure}[!hbt]
			\includegraphics [width=\linewidth, height=3cm]{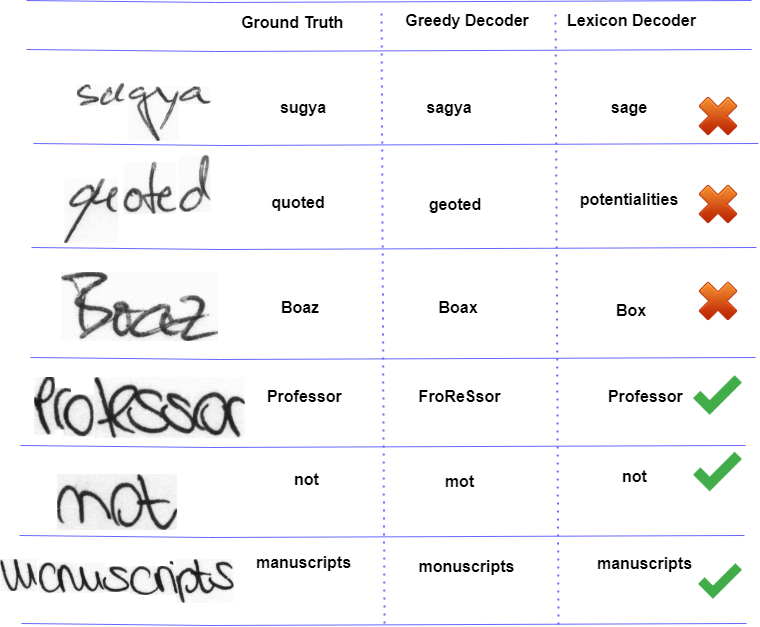}
			\caption {Some Positive and Negative outcomes are shown by applying the lexicon decoder as post processing step}
			\label{figure:fig-11}	
		\end{figure}
		
\begin{figure}
				\centering
				\subfigure[]{\includegraphics[width=0.5\linewidth, height=3cm]{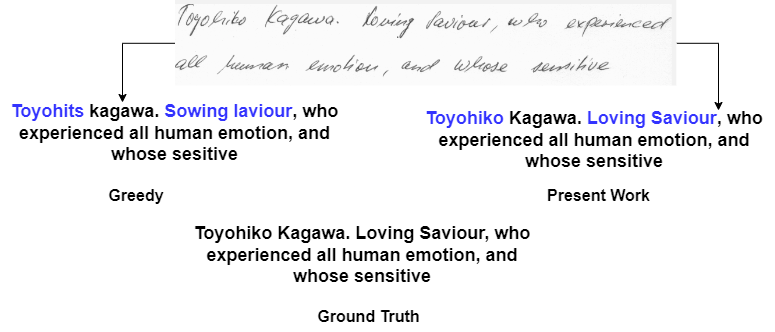}} 
				\subfigure[]{\includegraphics[width=0.49\linewidth, height=3cm]{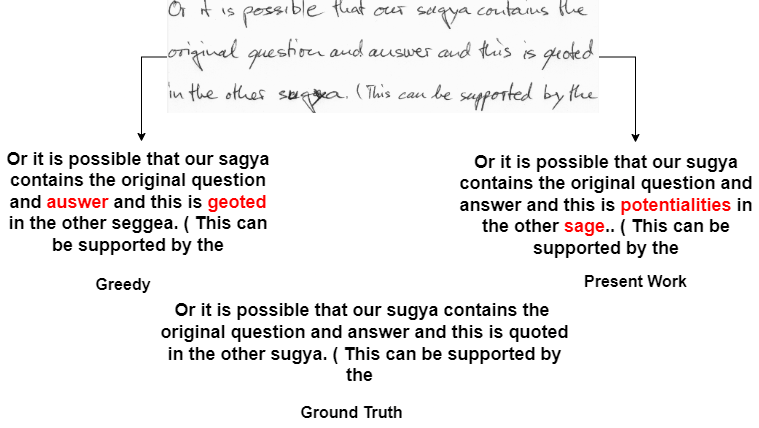}} 
				\caption{(a)If Greedy performs better, the WBS decoder made it even better  (b)Scenario in which WBS decoder reduces CER and WER in comparison with greedy decoder}
				\label{figure:fig-9-10}
			\end{figure}

		In this section, failure cases and best case scenarios  of the present  system have been discussed. The figure \ref{figure:fig-11} also depicts the performance of greedy and lexicon decoder on the IAM dataset. In the first three cases, lexicon decoder highly penalizes accuracy by predicting the wrong word. While in the last 3 cases lexicon decoder outperforms the greedy decoder. During the time of testing while applying the CTC WBS decoding algorithm, it internally builds a prefix tree and processes along the edges of the nodes of the prefix tree to predict the handwritten text. In worst cases, the path taken in the prefix tree is totally different from the actual ground truth. This results in a substantial increase in CER due to the prediction of totally a wrong word, in comparison with  the wrong prediction of one or two characters in words as in greedy decoder. As shown in Figure \ref{figure:fig-9-10},  the word \lq quoted\rq \  is predicted as \lq geoted\rq \ by the greedy decoder and \lq potentialities\rq \  by the WBS decoder. While word \lq sugya\rq \  is predicted as \lq seggea\rq \  by the greedy decoder and \lq sage\rq \  by the WBS decoder.
		
		\par While for best case scenarios, where our greedy decoder already performing with state-of-the-art accuracies ($<$1\% CER), the percentage improvement in such scenario is nearly perfect. For example, in Figure \ref{figure:fig-9-10} the word 'Toyohiko' is recognised as 'Toyohits' by the greedy decoder and  'Toyohiko' by the lexicon decoder.

			\vspace{-3em}
			\section{Conclusion}
			In this study, we propose a lexicon-based end-to-end paragraph recognition system lexiconNet. It is segmentation free and leverages the current state-of-the-art handwritten text recognition domain techniques. As a post processing step,  we have added a WBS decoder to the base model. We have reported substantial improvement in CER and WER on IAM, RIMES and READ-2016 datasets. We have also presented detailed analyses of best-case and worst-case observations due to the addition of a WBS decoder in the system which shows it is not always beneficial for having WBS as a decoder. This system is also able to recognize the complex layout of inclined lines. At present, this system is able to recognize the single-column layout of handwritten text only. To recognize a page in an end-to-end manner is a highly computationally intensive task. One of the future challenges may be making it less computationally extensive.

						\section*{Acknowledgment}
				 This research is funded by Government of India, University Grant Commission, under Senior Research Fellowship scheme. The authors acknowledge PRL’s supercomputing resource PARAM Vikram-1000 made  available for conducting the research reported in this paper
			
			\bibliographystyle{splncs04} 
			{ \footnotesize
				\bibliography{biblography}}

		\end{document}